\documentclass[runningheads]{llncs}
\usepackage{graphicx}
\usepackage{amsmath,amssymb,amsfonts}
\usepackage{multirow}
\usepackage{booktabs}
\usepackage{subfigure}
\usepackage{wrapfig}
\usepackage{xcolor}
\usepackage{bbding}
\usepackage{hyperref}

\authorrunning{Zhang et al.}
\begin{document}
	\title{Generator versus Segmentor: Pseudo-healthy Synthesis}
	\newcommand{\repeatthanks}{\textsuperscript{\thefootnote}}
	\author{
		Yunlong Zhang\inst{1} \thanks{equal contribution} \and
		Chenxin Li\inst{1} \repeatthanks \and
		Xin Lin\inst{1} \and
		Liyan Sun\inst{1} \and
		Yihong Zhuang\inst{1} \and
		Yue Huang \inst{1(}\Envelope\inst{)} \and
		Xinghao Ding\inst{1} \and
		Xiaoqing Liu\inst{2} \and
		Yizhou Yu\inst{2}
	}
	
	\institute{School of Informatics, Xiamen University, Xiamen, China \\ \email{yhuang2010@xmu.edu.cn} \and
		Deepwise Inc., Beijing, China}

	\maketitle              
	\begin{abstract}
		This paper investigates the problem of pseudo-healthy synthesis that is defined as synthesizing a subject-specific pathology-free image from a pathological one. Recent approaches based on Generative Adversarial Network (GAN) have been developed for this task. However, these methods will inevitably fall into the trade-off between preserving the subject-specific identity and generating healthy-like appearances. To overcome this challenge, we propose a novel adversarial training regime, Generator versus Segmentor (GVS), to alleviate this trade-off by a divide-and-conquer strategy. We further consider the deteriorating generalization performance of the segmentor throughout the training and develop a pixel-wise weighted loss by muting the well-transformed pixels to promote it. Moreover, we propose a new metric to measure how healthy the synthetic images look. The qualitative and quantitative experiments on the public dataset BraTS demonstrate that the proposed method outperforms the existing methods. Besides, we also certify the effectiveness of our method on datasets LiTS. Our implementation and pre-trained networks are publicly available at \url{https://github.com/Au3C2/Generator-Versus-Segmentor}.
		
	\end{abstract}
	\section{Introduction}
	\label{sec:introduction}
	Pseudo-healthy synthesis is defined as synthesizing a subject-specific pathology-free image from a pathological one \cite{xia2020pseudo}. Generating such images has been proven to be valuable for a variety of medical image analysis tasks \cite{xia2020pseudo}, such as segmentation \cite{bowles2017brain,ye2013modality,sun2020adversarial,andermatt2018pathology,Bowles2016PseudohealthyIS}, detection \cite{tsunoda2014pseudo}, and providing additional diagnostic information for pathological analysis \cite{baumgartner2018visual,sun2020adversarial}. By definition, \textit{a perfect pseudo-healthy image should maintain both healthiness (i.e., the pathological regions are indistinguishable from healthy ones in synthetic images) and subject identity (i.e., belonging to the same subject as the input).} Note that both of them are essential and indispensable. The importance of the former is self-explanatory, and the latter is also considerable since generating another healthy counterpart is meaningless.
	
	In this paper, we focus on promoting the pseudo-healthy synthesis from both above-mentioned aspects: healthiness and subject identity. \textit{The existing GAN-based methods attained promising results but still existed the trade-off between changing the entire appearance towards a healthy counterpart and keeping visual similarity.} Thus, we utilize a divide-and-conquer strategy to alleviate this trade-off. Concretely, we divide an image into healthy/pathological regions and apply the individual constraint for each of them. The first constraint is keeping visual consistency for healthy pixels before and after synthesis, and the second one is mapping pathological pixels into pixel-level healthy distribution (i.e., the distribution of healthy pixels).  Furthermore, to measure the distributional shift between healthy and pathological pixels, a segmentor is introduced into the adversarial training. 
	
	\noindent\textbf{Contributions.} Our contributions are summarized as three-fold.
	(1) We introduce a segmentor as the 'discriminator' by originality. The zero-sum game between it and the generator contributes to synthesize better pseudo-healthy images by alleviating the above-mentioned trade-off. 
	(2) We further consider the persistent degradation of generalization performance of the segmentor. To alleviate this issue, we propose a pixel-wise weighted loss by muting the well-transformed pixels.  
	(3) The only gold standard to measure the healthiness of synthetic images is the subjective assessment. However, it is time-consuming and costly, while being subject to inter- and intra-observer variability. Hence, it deviates from reproducibility. Inspired by the study of label noise, we propose a new metric to measure the healthiness.
	
	\noindent\textbf{Related work.} According to various clinical scenarios, a series of methods for pseudo-healthy synthesis were proposed  and mainly included the \textit{pathology-deficiency} (i.e., lacking pathological images in the training phase) \cite{chen2018unsupervised,sato2018primitive,schlegl2019f} and \textit{pathology-sufficiency} based methods (i.e., having plenty of pathological images in the training phase) \cite{baumgartner2018visual,sun2020adversarial,xia2020pseudo}. The pathology-deficiency based methods aimed to learn the normative distribution from plenty of healthy images. Concretely, they adopt the VAE \cite{sato2018primitive}, AAE \cite{chen2018unsupervised}, and GAN \cite{schlegl2019f} to reconstruct healthy images. In the testing phase, the out-of-distribution regions (i.e., lesions) cannot be reconstructed and were transformed into healthy-like ones. In contrast, the pathology-sufficiency based methods introduced pathological images along with \textit{image-level} \cite{baumgartner2018visual} or \textit{pixel-level} \cite{sun2020adversarial,xia2020pseudo} labeling. The existing methods aimed to translate pathological images into normal ones by the GAN. {\color{black}{Besides the adversarial loss that aligns the distributions between pathological and synthetic images, the VA-GAN \cite{baumgartner2018visual} introduced $\mathcal{L}_1$ loss to assure the visual consistency. In the process of applying Cycle-GAN \cite{zhu2017unpaired} to pseudo-healthy synthesis, the ANT-GAN proposed two improvements, which are the shortcut to simplify the optimization and the masked L2 loss to better preserve the normal regions. The PHS-GAN considered the one-to-many problem when applying the Cycle-GAN into pseudo-healthy synthesis. It disentangled the pathology information from what seems to be a healthy image part and then combined the disentangled information with the pseudo-healthy images to reconstruct the pathological images.}}
	
	\begin{figure*}[htbp]
		\centering
		\includegraphics[width=0.88\columnwidth]{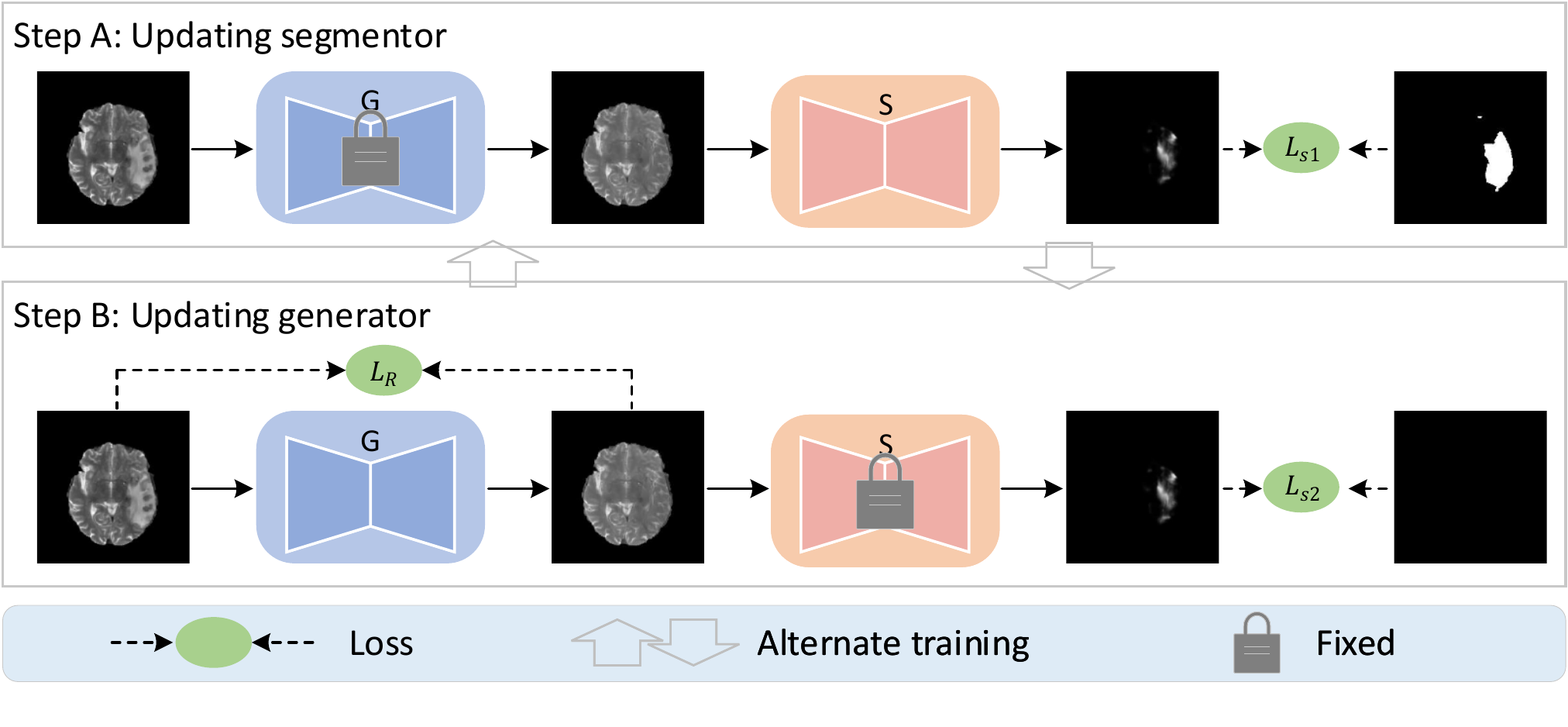}
		\caption{Training workflow. The model is optimized by iteratively alternating Step A and Step B. In Step A, we fix the generator $\mathbf{G}$ and update the segmentor $\mathbf{S}$ with $L_{s1}$. In Step B, we fix the segmentor $\mathbf{S}$ and update  the generator $\mathbf{G}$ with $L_{s2}+\lambda L_R$.}
		\label{fig1}
	\end{figure*}

	\section{Methods}
	\label{sec:Methods}
	In this section, the proposed GVS method for pathology-sufficiency pseudo-healthy synthesis with pixel-level labeling is introduced. Assume a set of pathological image $x_p$ and corresponding pixel-level lesion annotations $y_t$ are given.
	
	\subsection{Basic GVS flowchart}
	\label{subsection:flowchart}
	The training workflow of the proposed GVS is shown in Figure \ref{fig1}. The generator gradually synthesizes healthy-like images by iteratively alternating Step A and Step B. The specific steps are as follows.
	
	\noindent\textbf{Step A.} As shown in Figure \ref{fig1}, we fix the generator $\mathbf{G}$ and update the segmentor $\mathbf{S}$ to segment the lesions. The lesion annotation $y_t$ is adopted, and the loss is:
	\begin{equation}
		\mathcal{L}_{s1} = \mathcal{L}_{ce}(\mathbf{S}(\mathbf{G}(x_p)), y_t),
	\end{equation} 
	where $L_{ce}$ denotes the cross-entropy loss.
	
	\noindent\textbf{Step B.} In this step, we fix the segmentor $\mathbf{S}$ and update the generator $\mathbf{G}$, aiming to remove the lesions and preserve the identity of pathological images. Specifically, on the one hand, it is expected that the generator $\mathbf{G}$ can synthesize healthy-like appearances that do not contain lesions. Therefore, an adversarial loss is used:
	\begin{equation}
		\mathcal{L}_{s2} = \mathcal{L}_{ce}(\mathbf{S}(\mathbf{G}(x_p))), y_h),
	\end{equation} 
	where $y_h$ denotes the zero matrix with the same size as $y_t$. To deceive the segmentor, the generator further compensates the distributional difference between pathological and healthy regions. On the other hand, the synthetic images should be visually consistent with the pathological ones \cite{baumgartner2018visual,sun2020adversarial,xia2020pseudo}. Therefore, the generator $\mathbf{G}$ is trained with a residual loss:
	
	\begin{equation}
		\mathcal{L}_R = \mathcal{L}_{mse}(x_p,  \mathbf{G}(x_p)),
	\end{equation} 
	where $\mathcal{L}_{mse}$ denotes pixel-wise $\mathcal{L}_2$ loss.  The total training loss of  $\mathbf{G}$ is:
	\begin{equation}
		\mathcal{L}_{G} = \mathcal{L}_{s2} + \lambda \mathcal{L}_R,
	\end{equation} 
	where $\lambda$ denotes a hyperparameter that represents the trade-offs between the healthiness and identity, and it is subject to $\lambda > 0$.
	
	\subsection{Improved residual loss}
	\label{subsection:residual}
	This section further considers whether it is reasonable to set the same visual similarity in normal and pathological regions. Apparently, it is reasonable to keep visual consistency in normal regions between pathological and synthetic images. However, it is contradictory to remove lesions and keep pixel values within pathological regions. In order to alleviate this contradiction, we assume that potential normal tissues for lesion regions and normal tissues in the same pathological image have similar pixel values. Based on this, we improve the residual loss as follows:
	\begin{equation}
		\mathcal{L}_{R+} = \mathcal{L}_{mse}((1-y_t)\odot x_p,(1-y_t)\odot  \mathbf{G}(x_p)) + \lambda_1 \mathcal{L}_{mse}(y_t\odot \overline x_{pn},y_t\odot \mathbf{G}(x_p)),
	\end{equation} 
	where $\odot$ represents the pixel-wise multiplication, and $\overline x_{pn}$ denotes a matrix filled with the average value of normal tissue in the same image $x_p$ and has the same size as $x_p$; $\lambda_1$ denotes a hyperparameter that controls the power of visual consistency in the lesion regions, and $0<\lambda_1<1$ since the potential normal tissues are close but not equal to the average value of normal tissues.
	
	\subsection{Training a segmentor with strong generalization ability}
	\label{subsection:segmentor}
	The generalization ability of segmentor is further considered. During the training, the pathological regions are gradually transformed into healthy-like ones. As shown in Figure \ref{fig2}(b), the major part of the lesion region has been well transformed, so these pixels should be labeled as 'healthy.' However, the basic GVS still views all the pixels in this region as lesions. In other words, the well-transformed parts are forced to fit the false label, which weakens the generalization capacity of neural networks \cite{zhang2016understanding}. 
	As shown in Figure \ref{fig2}(b),  we observe that predictions of segmentor substantially deviate from  labels, which suggests the poor generalization of the segmentor.  To meet this challenge, we present a novel pixel-level weighted cross-entropy loss for lesion segmentation. The difference map between pathological and synthetic images can be used as an indicator to measure the transformation degree:
	\begin{figure}[htbp]
		\centering
		\includegraphics[width=0.88\columnwidth]{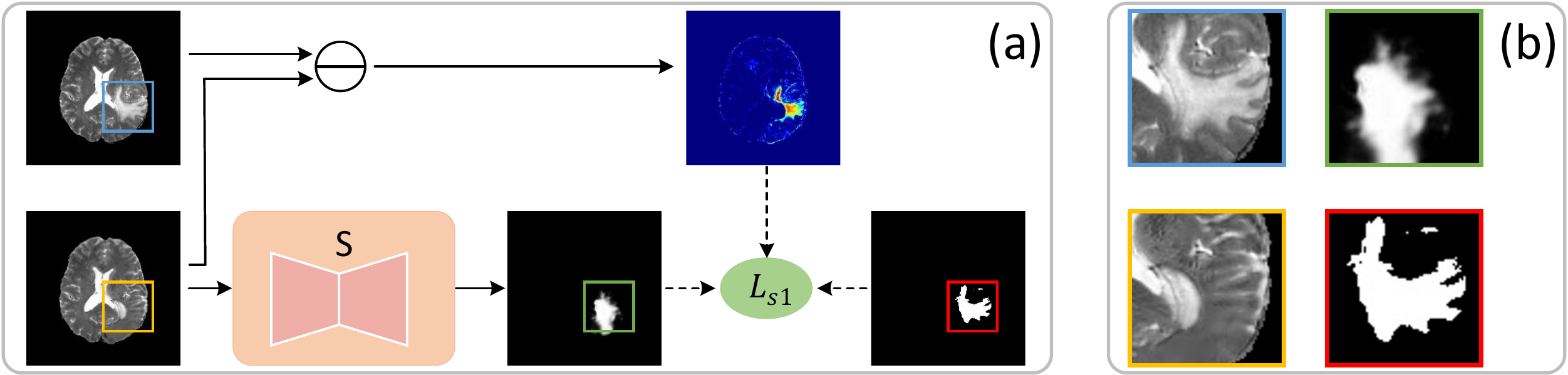}
		\caption{(a) Framework of the pixel-level weighted cross-entropy loss. (b) The blue, yellow, green, and red boxes denote  zoom-in views corresponding to the pathological image, synthetic image, segmentation prediction, and lesion annotation, respectively.}
		\label{fig2}
	\end{figure}
	\begin{equation}
		\mathcal{L}_{wce} = \frac{1}{N}\sum_{i=1}^{N}w(i)y_{t}(i)log(\mathbf{S}(\mathbf{G}(x_p))(i)),
	\end{equation} 
	where $N$ denotes the number of pixels. The weight $w$ associated with difference maps are defined as:
	
	\begin{equation}
		w =  \left\{
		\begin{array}{rcl}
			0.1,       &      & 1 - m< 0.1,\\
			1 - m,       &      & \text{Otherwise},
		\end{array} \right.
	\end{equation} 
	where $m = \text{Normalization}(x_p - \mathbf{G}(x_p))$ denotes the normalized difference map. 
	In this work, $w[w < 0.1] = 0.1$ because the minimum value does not represent perfect-transformation, and it is necessary to keep a subtle penalty.
	
	The complete GVS is developed by replacing $\mathcal{L}_R$ with $\mathcal{L}_{R+}$, and then replacing $\mathcal{L}_{s1}$  with $\mathcal{L}_{wce}$. Please note that the GVS refers to the complete GVS in the next subsections.
	
	\section{Experiments}
	\subsection{Implementation details}
	\noindent\textbf{Data.} {\color{black}{Our method is mainly evaluated on BraTS2019 dataset \cite{menze2014multimodal,bakas2017advancing},}} which contains 259 GBM (i.e., glioblastoma) and 76 LGG (i.e., lower-grade glioma) volumes. Here, we utilize the T2-weighted volumes of GBM, and they are split into training (234 volumes) and test sets (25 volumes). {\color{black}{To verify the potential to different modalities and organs, we also present some visual results on the LiTS \cite{bilic2019liver}, which}} contains 131 CT scans of the liver. The slice resolution is $512\times512$. The datasets are divided into training (118 scans) and test sets (13 scans). The intensity of images is rescaled to the range of $[0,1]$.
	
	\noindent\textbf{Network.} Both the generator and the segmentor adopt the 2D U-Net architectures \cite{ronneberger2015u}, which consist of an encoder-decoder architecture with symmetric skip connections. They both downsample and upsample four times and adapt the bilinear upsampling method. Furthermore, unlike the generator, the segmentor contains the instance normalization and softmax layer.
	
	\noindent\textbf{Training details.} The proposed method is implemented on Pytorch and an NVIDIA TITAN XP GPU. We use Adam optimizer, with an initial learning rate of $0.001$ and decrease it by 0.1 after $0.8*total\_epoch$.  The $total\_epoch$ is set to $20$. The batch size is set 8 for the BraTS and 4 for the LiTS. Lastly, $\lambda$ and $\lambda_1$ are set to $1.0$ and $0.1$.

	\subsection{Evaluation metrics}
	In this section, two metrics, $\mathbb{S}_{dice}$ and $iD$, are introduced to evaluate the healthiness and identity, respectively.
	
	Zhang et al. \cite{zhang2016understanding} revealed an interesting phenomenon that convergence time on the false/noisy labels increases by a constant factor compared with that on the true labels. Similar to this, aligning well-transformed pixels (i.e., these pixels can be viewed as healthy ones) and lesion annotations is counterfactual and hampers the convergence. Thus, how healthy the synthetic images look is negatively related to the convergence time. Inspired by this, we present a new metric to assess the healthiness, which is defined as the accumulated dice score throughout the training process: $\mathbb{S}_{dice} = \sum_{e=1}^{epochs} dice_e$, where $epochs$ denotes the number of training epochs, and $dice_e$ denotes the dice evaluated on the training data at the $e$-th epoch. 
	
	\begin{wrapfigure}[14]{r}{0.4\textwidth}
		\centering
		\vspace{-1.5\intextsep}
		\hspace*{-.75\columnsep}\includegraphics[width=0.4\textwidth]{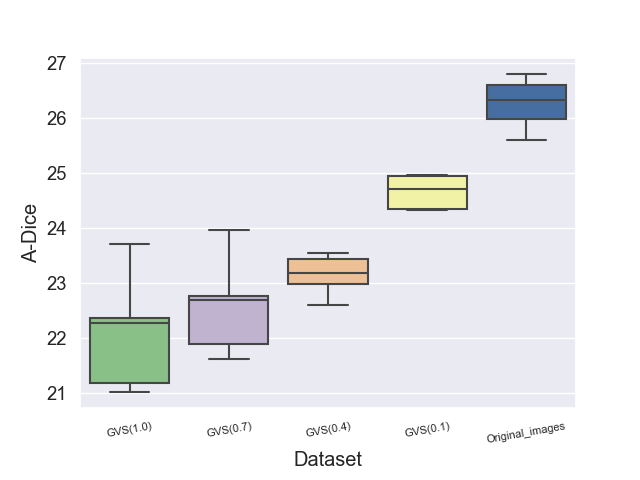}
		\caption{The $\mathbb{S}_{dice}$ evaluated on different synthetic images. The GVS(1.0) represents the synthetic images generated by the GVS that is trained on $100\%$ of training data.}
		\label{fig3}
	\end{wrapfigure}
	
	Here, whether the $\mathbb{S}_{dice}$ can correctly assess the healthiness of synthetic images is testified. To this end, we calculate the $\mathbb{S}_{dice}$ of GVS(0.1), GVS(0.4), GVS(0.7), GVS(1.0), and original images. Note that the GVS(a) denotes the synthetic images generated by the GVS trained on $a\%$ training data. Generally, the GVS(a) should have more healthy appearances than the GVS(b) when $a > b$. The results show that the $\mathbb{S}_{dice}$ can correctly reflect the healthiness in order. In addition, we also discover that the $\mathbb{S}_{dice}$ is unstable in the small value area, which may result in false results when synthetic images have similar healthy appearances. One possible reason for this situation is that the random parameter initialization may lead to different results at each trial.
	\begin{figure}[!h]
		\centering
		\includegraphics[width=\columnwidth]{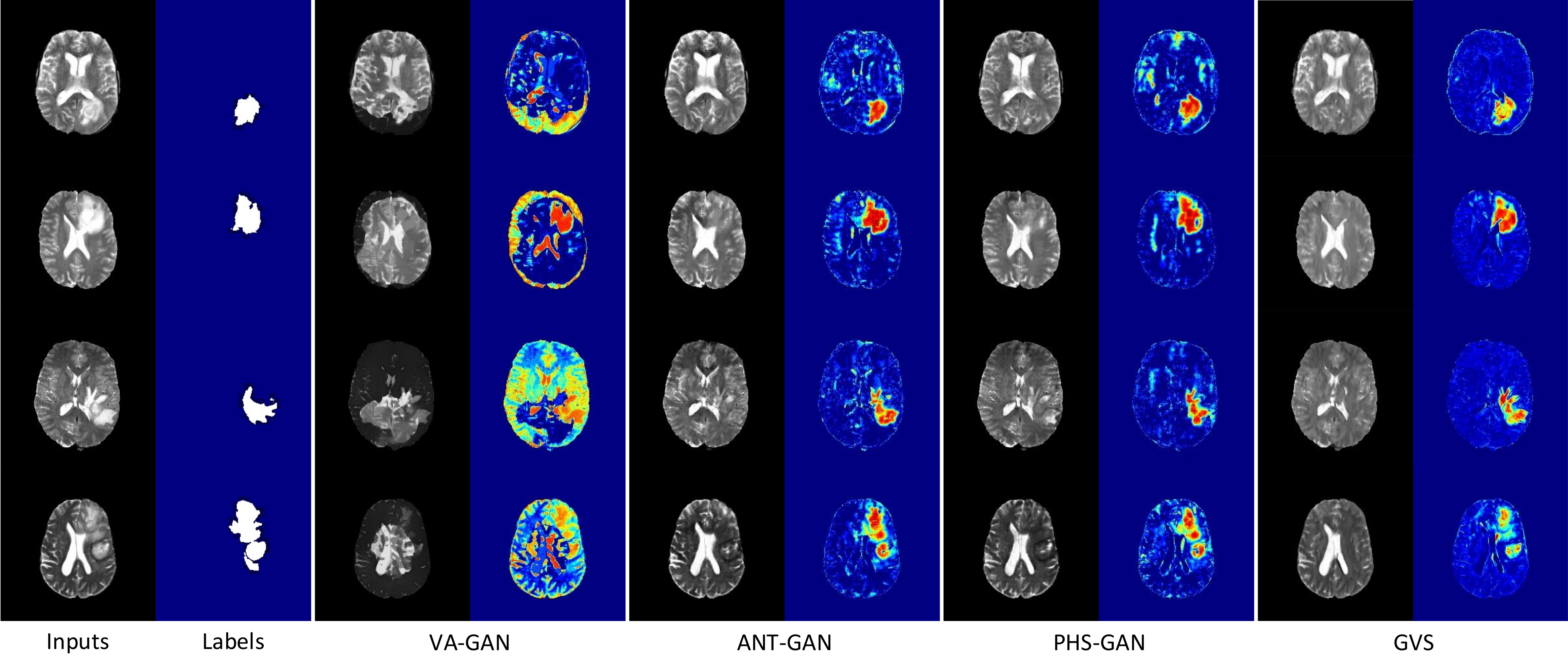}
		\caption{The qualitative results of synthetic images. The first and second columns show inputs and corresponding labels, respectively. The next eight columns show the synthetic images and difference maps of the VA-GAN, ANT-GAN, PHS-GAN, and GVS.}
		\label{fig4}
	\end{figure}
	
	Following Xia et. \cite{xia2020pseudo}, the identity is expressed as the structural similarity calculated on normal pixels, which is defined as $iD = \text{MS-SSIM}[(1-y_t) \odot \mathbf{G}(x_p), (1-y_t) \odot x_p]$, where the $\odot$ denotes the pixel-wise multiplication, and the $\text{MS-SSIM}()$ representes a masked Multi-Scale Structural Similarity Index \cite{wang2003multiscale}.

	\subsection{Comparisons with other methods}
	We compare our GVS with existing pathology-deficiency based methods, including the VA-GAN \cite{baumgartner2018visual}, ANT-GAN \cite{sun2020adversarial}, and PHS-GAN \cite{xia2020pseudo}. The VA-GAN\footnote{https://github.com/baumgach/vagan-code} and PHS-GAN\footnote{https://github.com/xiat0616/pseudo-healthy-synthesis} are implemented using their official codes, and the ANT-GAN is implemented on the code provided by the authors. Note that the VA-GAN uses \textit{image-level} labels, and the ANT-GAN and PHS-GAN utilize \textit{pixel-level} labels. Our experimental setting keeps \textit{consistent} with the last two. Next, we analyze the performance of all methods qualitatively and quantitatively. 
	
	\begin{table}
		\centering
		\caption{Evaluation results of the VA-GAN, PHS-GAN, ANT-GAN, GVS and its variants, as well as the baseline. The performance of the baseline was evaluated on original images. The average values and standard deviations of the evaluation results were calculated over three runs. The best mean value for each metric is shown in bold.}	
		\begin{tabular}{cccccccc}
			\toprule
			~ & VA-GAN & PHS-GAN & ANT-GAN & GVS & 	w/o $\mathcal{L}_{R+}$ & w/o $\mathcal{L}_{wce}$ & Baseline \\
			\hline
			$iD \uparrow$ & $0.74_{0.03}$ & $0.97_{0.03}$ & $0.96_{0.02}$ & $\textbf{0.99}_{0.01}$ & $\textbf{0.99}_{0.01}$ & $\textbf{0.99}_{0.01}$ & $1.00_{0.00}$ \\
			$\mathbb{S}_{dice} \downarrow$ & $-$ & $23.32_{0.45}$ & $23.11_{0.41}$ & $\textbf{21.75}_{0.13}$ & $22.11_{0.14}$ & $23.66_{0.15}$ & $26.23_{0.17}$ \\
			\bottomrule
		\end{tabular}
		\label{table1}
	\end{table}

	Qualitative results are shown in Figure \ref{fig4}. We first analyze the identity by comparing the reconstruction performance in the normal regions. The proposed method achieves higher reconstruction quality than the other methods (better see the difference maps). For example, compared with the existing methods, our method better preserves the high-frequency details (i.e., edges). Overall, the VA-GAN could not keep the subject identity and loses a part of lesion regions in some cases. The PHS-GAN and ANT-GAN preserve the major normal regions but lose some details. Among all the methods, the proposed method achieves the best subject identity. Then, we further analyze the healthiness, which can be judged by comparing whether pathological and normal regions are harmonious. The synthetic images generated by VA-GAN are visually unhealthy due to the poor reconstruction. The PHS-GAN and ANT-GAN remove most of the lesions, but some artifacts still remain. The GVS achieves the best healthiness due to indistinguishable pathological and normal regions.
	
	\begin{figure}[!h]
		\centering
		\subfigure[]{
			\includegraphics[width=0.3\columnwidth]{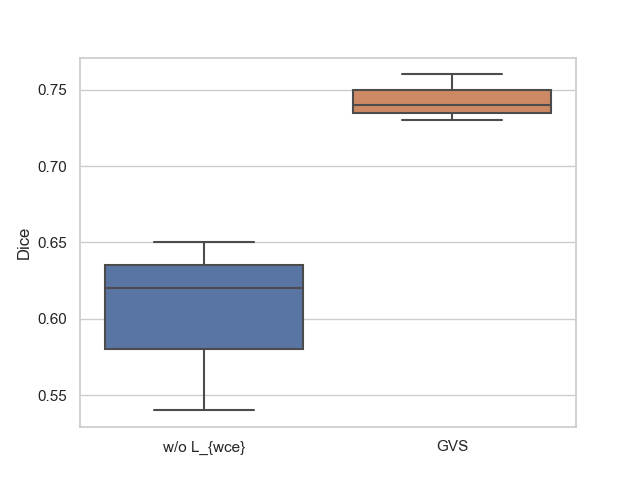}
			\label{fig52}}
		\subfigure[]{
			\includegraphics[width=0.55\columnwidth]{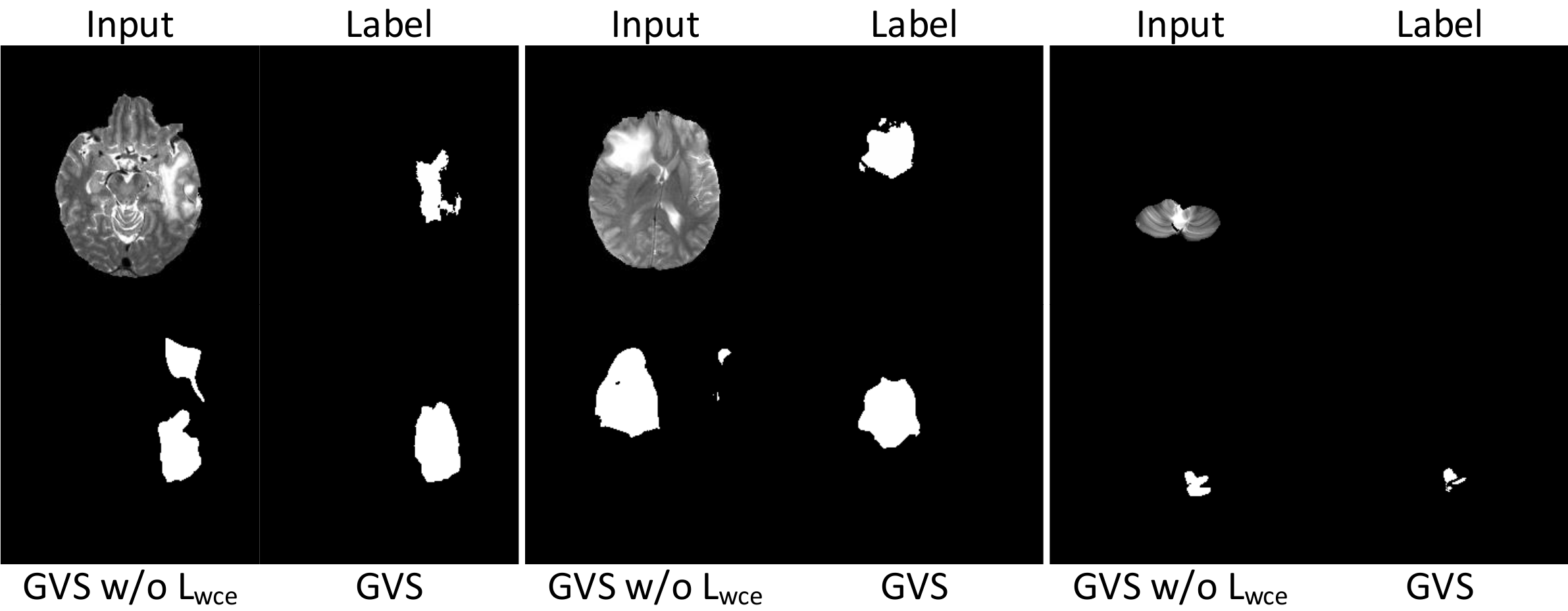}
			\label{fig51}}
		\caption{(a)  Generalization ability of the segmentors in the GVS and GVS w/o $\mathcal{L}_{wce}$. (b) Visual comparison of synthetic images of the GVS and GVS w/o $\mathcal{L}_{wce}$. Three blocks show three examples. Each block contain the input, label, predictions of the GVS without $\mathcal{L}_{R+}$ and GVS.
		}\label{fig5}
	\end{figure}

	We report quantitative results in Table \ref{table1}. The first metric, $iD$, is used to assess the identity. Our GVS achieves a $iD$ value of 0.99, outperforming the second place, PHS-GAN, by 0.02, which is an evident improvement when the SSIM is large. The proposed $\mathbb{S}_{dice}$ is used to assess healthiness. Since the VA-GAN can not reconstruct the normal tissue, as shown in the third column in Figure \ref{fig4}), its $\mathbb{S}_{dice}$ value is meaningless and not considered. In addition, compared with baseline (i.e., original images), the $\mathbb{S}_{dice}$ of the ANT-GAN and PHS-GAN  decline from 26.23 to 23.32 and 23.11, respectively. 
	The proposed method further improves the $\mathbb{S}_{dice}$ to 21.75.

	\subsection{Ablation study}

	To verify the claim that the $\mathcal{L}_{wce}$ can alleviate the poor generalization of the segmentor, we respectively calculate the segmentation performance of the segmentors trained by the GVS and GVS w/o $\mathcal{L}_{wce}$. As shown in Figure \ref{fig52}, the GVS achieves a higher average dice score and lower variance compared to the GVS w/o $\mathcal{L}_{wce}$, which confirms that the $\mathcal{L}_{wce}$ could effectively improve the generalization ability of the segmentor. Similar conclusions are also derived from the visual examples in Figure \ref{fig51}. Predictions of segmentor trained by GVS w/o $\mathcal{L}_{wce}$ deviate from the labels severely. After adding the $\mathcal{L}_{wce}$, the predictions are more accurate. Note that only relying on the $\mathcal{L}_{wce}$ cannot solve the generalization problem of the segmentor entirely (see the third example in Figure \ref{fig51}). Hence, this problem needs further exploration in the future. Furthermore, benefiting from the better generalization of the segmentor, the healthiness attains further improvement ($23.66\rightarrow 21.75$ in Table \ref{table1}). We also conduct the ablation study on the improved residual loss, and results are shown in Table \ref{table1}. We observe that the $\mathbb{S}_{dice}$ increases from $21.75$ to $22.11$ after replacing $\mathcal{L}_{R}$ with $\mathcal{L}_{R+}$.
	
	\begin{figure}[!h]
		\centering
		\includegraphics[width=\columnwidth]{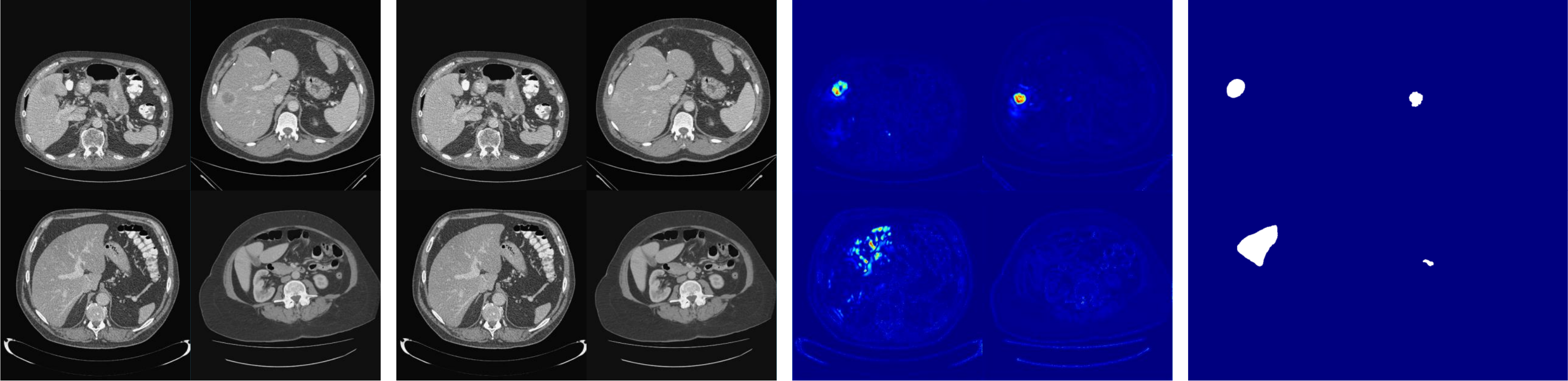}
		\caption{The visual results on the LiTS dataset. The blocks from left to right represent the pathological images, synthetic images, difference maps, and lesion annotations.}
		\label{fig7}
	\end{figure}

	\subsection{Results on LiTS dataset}
	The proposed GVS is also evaluated on a public CT dataset, LiTS. The results show that the proposed method can maintain the identity and transform the high-contrast lesions well, as shown in the top row in Figure \ref{fig7}, However, the synthetic results of low-contrast images still exist subtle artifacts in the pathological regions, as shown in the bottom row in Figure \ref{fig7}. We conjecture that the reason may be that the segmentor cannot accurately detect low-contrast lesions, which further results in poor transformation.
	
	\section{Conclusions}
	This paper proposes an adversarial training framework by iteratively training the generator and segmentor to synthesize pseudo-healthy images. Also, taking the rationality of residual loss and the generalization ability of the segmentor into account, the improved residual loss and pixel-wise weighted cross-entropy loss are introduced.  In experiments, the effectiveness of the proposed scheme is verified on two public datasets, BraTS and LiTS.
	
	\noindent\textbf{Limitations and Future Work.} One limitation of the proposed GVS is requiring densely labeled annotations. In clinical application, huge amounts of accurate segmentation labels are hardly available. Hence, it is necessary to relax the demand for accurate pixel-level annotations in the next step.  Another concern is the instability of the proposed $\mathbb{S}_{dice}$. In the future, we plan to improve this by further exploiting the more characteristics of noise labels.

	\noindent\textbf{Acknowledgements.} This work was supported in part by National Key Research and Development Program of China (No. 2019YFC0118101), in part by National Natural Science Foundation of China under Grants U19B2031, 61971369, in part by Fundamental Research Funds for the Central Universities 20720200003,  in part by the Science and Technology Key Project of Fujian Province, China (No. 2019HZ020009).
	
	\bibliographystyle{splncs04}
	\bibliography{GVS}
	
\end{document}